\documentclass{article}

\PassOptionsToPackage{numbers,compress}{natbib}

\usepackage[dblblindworkshop, final]{neurips_2026}
\workshoptitle{New in Machine Learning (NewInML)}

\usepackage[utf8]{inputenc}
\usepackage[T1]{fontenc}
\usepackage{graphicx}       % figures, \resizebox
\usepackage{amsmath}        % align, cases
\usepackage{amssymb}        % \mathbb
\usepackage{booktabs}       % \toprule etc.
\usepackage{xcolor}         % colors
\usepackage{enumitem}       % compact lists
\usepackage{url}
\usepackage{microtype}
\usepackage{hyperref}
\usepackage[capitalize]{cleveref}  % \Cref -- must load after hyperref

\title{2D Gum-Net: Unsupervised Dense Elastic Image Registration via Symmetric Correlation and Spectral Pooling}

\author{%
  Aarya Bhave \\
  Google \\
  \texttt{aaryabhave@google.com} \\
  \And
  Anushka Ajay Bhave \\
  Google \\
  \texttt{nushiva@google.com} \\
  \And
  Akanksha Narula \\
  Google \\
  \texttt{akankshanarula@google.com} \\
  \AND
  Marius Sebastian Shokoui \\
  Technical University of Munich\\
  \texttt{marius.shokoui@tum.de} \\
  \And
  Alina Hafner \\
  Technical University of Munich \\
  \texttt{ alina.hafner@tum.de} \\
}

\begin{document}

\maketitle

\begin{abstract}
Dense non-rigid registration, the recovery of a pixel-wise deformation field that warps one image into geometric agreement with another, is fundamental across medical imaging, structural biology, and correspondence estimation, yet learning-based approaches face a severe bottleneck: ground-truth deformation fields are notoriously difficult to obtain for real data. We present 2D Gum-Net, an end-to-end architecture that predicts dense elastic deformation fields fully unsupervised. We transfer the Geometric Unsupervised Matching Network (Gum-Net), originally developed for 3D subtomogram alignment in cryo-electron tomography, to 2D elastic registration: all three modules are redesigned in 2D, and the rigid transformation head is replaced by a Dense Spatial Transformer that regresses a sparse control-point grid, lifted to a full-resolution field $\mathbf{\Phi}$ by bicubic interpolation. Downsampling uses DCT spectral pooling, implemented as frozen-basis matrix multiplications (${\approx}50\times$ GPU speedup), preserving the fine, quasi-periodic structure that max and average pooling destroy. We evaluate on contact-based fingerprints, a challenging testbed for this task: locally self-similar ridge patterns make correspondence maximally ambiguous, skin pressed against a sensor generates authentic large-displacement elastic distortion, and minutiae act as precise landmarks through which an independent matcher externally verifies the topological correctness of predicted fields. Trained purely on synthetic impressions and evaluated zero-shot on real benchmarks, our method improves structural correlation by up to 15.7\% on FVC2004 and raises genuine match scores under an independent matcher while leaving imposter scores essentially unchanged. This suggests that the predicted deformations restore true structure without fabricating correspondences.
\end{abstract}

\section{Introduction}
\label{sec:intro}

\begin{figure}[t]
  \centering
  \includegraphics[width=0.82\linewidth]{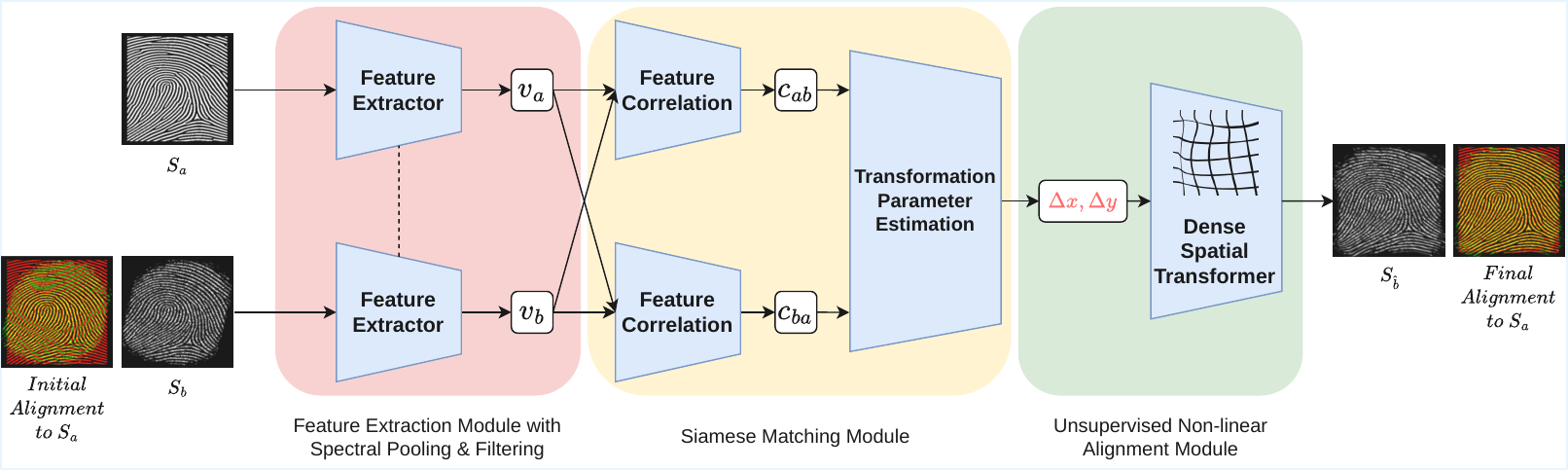}
  \caption{2D Gum-Net pipeline. The model is unsupervised and feed-forward. It inputs a geometrically distorted image $S_b$ and a reference image $S_a$ (here a fingerprint impression and its template) and outputs the un-warped image $S_{\hat{b}}$ in geometric agreement with $S_a$. The parameters of this transformation are denoted by $(\Delta x, \Delta y)$. The dashed line denotes that weights are shared between the two feature extractors.}
  \label{fig:gumnet_pipeline}
\end{figure}

Dense non-rigid registration is the problem of recovering a pixel-wise deformation field $\mathbf{\Phi}$ such that a moving image $S_b$, resampled through the field, comes into geometric agreement with a reference image $S_a$. Formally, one seeks
\begin{equation}
\mathbf{\Phi}^{*} = \arg\min_{\mathbf{\Phi}} \; \mathcal{D}\!\left(S_a,\; S_b \circ \mathbf{\Phi}\right) + \lambda\,\mathcal{R}(\mathbf{\Phi}),
\label{eq:registration}
\end{equation}
where $\mathcal{D}$ measures structural dissimilarity and $\mathcal{R}$ regularizes the field toward physically plausible, smooth deformations. This formulation underlies deformable medical image registration, dense correspondence estimation in natural images, and structural alignment in biology \cite{balakrishnan_voxelmorph_2019, rocco_convolutional_2017, schmidt_self_2016, zeng_gum-net_2020}. Classical methods solve \Cref{eq:registration} by iterative per-pair optimization, which is accurate but slow and sensitive to initialization. Learning-based methods amortize this optimization into a feed-forward predictor of $\mathbf{\Phi}$ \cite{rocco_convolutional_2017, si_dense_2017, cui_dense_2019, guan_regression_2023}, but most require supervised ground-truth deformation fields, which for real data essentially never exist and must be simulated \cite{guan_regression_2023, grosz_c2cl_2021}. Unsupervised dense registration, which trains from image pairs alone using the similarity functional itself as the learning signal, removes this bottleneck \cite{balakrishnan_voxelmorph_2019}. A second, less discussed obstacle is architectural: correspondence information lives in fine image structure, yet max pooling retains only localized maximal responses and average pooling attenuates and blurs local structure \cite{rippel_spectral_2015, zhang_hartley_2018, smith_discrete_2018}, degrading the very signal a registration network must correlate.

In this work we transfer an unsupervised registration paradigm across both domain and dimensionality. The Geometric Unsupervised Matching Network (Gum-Net) of Zeng et al.\ \cite{zeng_gum-net_2020} aligns 3D subtomograms in cryogenic electron tomography (cryo-ET) without any labels, reducing alignment error by 40--50\% while accelerating the overall process 70--110$\times$. Its mechanism, dense symmetric correlation of deep feature maps followed by transformation regression, is largely modality-agnostic \cite{rocco_convolutional_2017, zeng_gum-net_2020}. We hypothesize that the same mechanism, redesigned in two dimensions, solves general 2D elastic registration. Our adaptation (\Cref{fig:gumnet_pipeline}) differs from the original in two fundamental ways: all three modules are rebuilt for 2D inputs, and, more consequentially, the rigid transformation head is replaced with a Dense Spatial Transformer, which regresses a sparse control-point grid of $2G^2$ local displacements lifted to a full-resolution dense field $\mathbf{\Phi}$ by bicubic interpolation. This changes the transformation class from rigid motions to a smooth, expressive family of elastic warps, keeping the field smooth by construction and the parameter count small.

We evaluate this framework on contact-based fingerprint alignment, and we argue that fingerprints are a deliberately demanding testbed for dense elastic registration rather than merely an application. First, ridge patterns are quasi-periodic and locally self-similar: sliding a patch by one ridge wavelength produces a near-perfect local match that is globally wrong, so local matching faces an extreme aperture problem and a registration network succeeds only by integrating long-range spatial context. Second, the deformations are authentic and free: elastic skin pressed and twisted against a rigid sensor produces genuine large-displacement, non-linear warps in every acquisition \cite{bazen_fingerprint_2003, maltoni_handbook_2009, gao_study_2012}, on which rigid and affine models demonstrably fail. Third, the domain supplies an \emph{independent verification oracle}. Registration evaluation is usually circular, since alignment quality is measured with the same class of similarity functional that was optimized. Fingerprint minutiae, by contrast, are precise point landmarks consumed by an external matching algorithm: a field that restores true structure raises genuine match scores while leaving imposter scores unchanged, whereas a field that hallucinates or folds structure is immediately penalized. Fourth, the correspondence signal is concentrated in a narrow fine-scale band (the ridge frequency), so fingerprints isolate the causal effect of information-preserving downsampling: pooling that destroys the ridge band destroys registration. Finally, procedural synthetic generators provide unlimited template--impression training pairs, leaving real benchmarks untouched for zero-shot transfer evaluation.

In summary, the primary contributions of this work are as follows:
\begin{itemize}[leftmargin=1.4em,itemsep=1pt,topsep=2pt]
    \item \textbf{Cross-domain transfer of unsupervised registration:} We transfer the Gum-Net paradigm from volumetric cryo-ET alignment to 2D elastic registration, redesigning all three modules and replacing rigid regression with a Dense Spatial Transformer that predicts a control-point-parameterized dense field $\mathbf{\Phi}$, trained fully unsupervised, circumventing the deformation ground-truth bottleneck.

    \item \textbf{Information-preserving spectral downsampling:} We implement DCT spectral pooling as frozen-basis matrix multiplications (${\approx}10\times$ GPU speedup over FFT-based extraction) and ablate it against max and average pooling to isolate its causal effect on registration quality.

    \item \textbf{Externally validated evaluation:} We evaluate predicted fields with structural correlation and through an independent minutiae matcher acting as a topological-correctness oracle, showing that alignment raises genuine similarity while leaving imposter similarity unchanged.
\end{itemize}
We deliberately evaluate depth-first on a single demanding domain; downstream minutiae-matching results serve as a probe of topological correctness and genuine-score separability, not as a benchmark of a complete recognition system.

\section{Related work}
\label{sec:relwork}

\subsection{Dense non-rigid registration and geometric matching}
Learned dense correspondence is commonly built on feature correlation and cost-volume constructions that score every spatial location of one feature map against every location of the other, from which displacement fields are regressed \cite{rocco_convolutional_2017, schmidt_self_2016}. Spatial Transformer Networks \cite{jaderberg_stn_2015} established the differentiable-resampling machinery for training end-to-end through a predicted warp; our Dense Spatial Transformer extends this from parametric transformations to a control-point-parameterized dense field. In medical imaging, VoxelMorph \cite{balakrishnan_voxelmorph_2019} showed dense registration networks can be trained \emph{unsupervised}, with an image-similarity loss plus smoothness penalty replacing deformation ground truth; in effect, classical intensity-based registration amortized into a feed-forward predictor. In the fingerprint domain, dense registration and displacement-field regression have been studied classically and with deep networks \cite{si_dense_2017, cui_dense_2019, guan_regression_2023, guan_phase_2024}, but predominantly supervised, trained on simulated or annotated distortion fields. Fully unsupervised estimation of dense non-linear deformation fields remains comparatively unexplored outside the medical domain, and is the setting this work addresses.

\subsection{Elastic registration in fingerprint biometrics}
Our testbed has its own registration literature. Traditional fingerprint systems are minutiae-based: ridge endings and bifurcations are matched after estimating a geometric transformation between point sets \cite{maltoni_handbook_2009}, typically rigid or locally affine, via Hough-style voting \cite{ratha_realtime_1996} and RANSAC refinement with local structural descriptors \cite{feng_combining_2008, cappelli_mcc_2010}. These approaches assume near-rigid geometry and depend on reliable minutiae extraction, which degrades under distortion; since real impressions exhibit non-linear stretching and torque, elastic deformation remains a primary source of false rejections \cite{bazen_fingerprint_2003}. Classical elastic registration addressed this directly: Bazen and Gerez fit thin-plate splines to minutiae correspondences \cite{bazen_fingerprint_2003}, while Ross, Dass, and Jain estimate average deformation models \cite{ross_deformable_2005} and non-linear warps from ridge curve correspondences \cite{ross_warping_2006}. Modern learned pipelines such as C2CL \cite{grosz_c2cl_2021} and contrastive cross-domain approaches \cite{pandey_ridgeformer_2024} retain explicit alignment modules (e.g., TPS-based rectification), but optimize them through pretrained identity networks or supervised metric learning.

Our method is best understood as the \emph{learned, amortized analogue} of this classical elastic lineage: a control-point grid lifted by smooth interpolation is structurally a free-form deformation model, a close cousin of TPS. But where classical methods solve a per-pair optimization from explicitly extracted correspondences, inheriting the fragility of the extraction step, our network amortizes the inference into a single forward pass from raw pixels, learning the correspondence function itself without supervision.

\subsection{Spectral representations and pooling}
Frequency-domain techniques have long served the testbed domain for enhancement, translation-invariant matching, and compact descriptors \cite{jirachaweng_fingerprint_2007, al-ani_biometrics_2013, ito_fingerprint_2004, ouyang_fingerprint_2006, xu_spectral_2010}, but for representation or rigid alignment rather than for learning non-rigid correspondence. Separately, spectral pooling (subsampling by truncating high-frequency transform coefficients) preserves substantially more spatial information per parameter than max or average pooling, with DFT \cite{rippel_spectral_2015}, DCT \cite{smith_discrete_2018}, and Hartley \cite{zhang_hartley_2018} variants. This property is not domain-specific: it is a principled alternative to spatial subsampling whenever deformation must be estimated from fine structure, which is exactly the regime our testbed maximizes.

\subsection{Unsupervised learning in the testbed domain}
Unsupervised learning has been explored for fingerprints through RBMs, autoencoders, and GAN-based enhancement \cite{sahasrabudhe_learning_2013, karabulut_cyclegan_2020, hou_unsupervised_2021}, showing that structural priors can be learned without annotation, while deep matchers, whether CNN-based \cite{tang_fingernet_2017, nguyen_minutiaenet_2018, engelsma_deepprint_2019, pan_flare_2024} or transformer-based \cite{maity_latentprintformer_2024, grosz_afrnet_2024}, learn identity-discriminative features without explicitly modeling elastic deformation, which remains a major cause of matching errors. In contrast, we learn the deformation field itself, densely and without labels.

\section{Method}
\label{sec:method}
Our model is shown in \Cref{fig:gumnet_pipeline}. It takes two grayscale images: the reference (fixed) image $S_a$ and the moving image $S_b$ to be registered onto it. In our evaluation domain these are a fingerprint master template and an impression of the same finger, pairs that synthetic generators produce natively (\Cref{sec:experiments}). Weight-shared feature extractors produce feature maps $v_a$ and $v_b$, which a Siamese matching module converts into two correlation maps: at spatial coordinate $(i,j)$, $c_{ab}$ holds the similarity between $v_a$ at $(i,j)$ and all features of $v_b$, with $c_{ba}$ defined symmetrically. Both maps are processed by identical network branches, flattened, and concatenated to estimate the transformation parameters. The spatial transformation is realized by a dense deformation field $\mathbf{\Phi} \in \mathbb{R}^{H \times W \times 2}$; however, to ensure smoothness and reduce complexity, we do not regress this field directly. Instead it is parameterized by a sparse control-point grid $\mathbf{P} \in \mathbb{R}^{2 \times G \times G}$: for each control point $P_i$, $i \in \{1, ..., G^2\}$, the network regresses a localized translation $\Delta \mathbf{p}_i = (\Delta x_i, \Delta y_i)$, and the resulting $2G^2$ parameters are upsampled via bicubic interpolation to the full-resolution field $\mathbf{\Phi}$. The objective is to find the field that maximizes structural overlap between $S_a$ and the warped $S_{\hat{b}}$, completely unsupervised: a soft Dice alignment loss \cite{wang_dice_2023} plus a smoothness regularizer (\Cref{subsec:objective}) is back-propagated through the differentiable warp.

\subsection{Feature extraction module} \label{subsec: feature_extraction_module}
Popular feature-extraction backbones such as DenseNet \cite{huang_densenet_2017}, InceptionNet \cite{szegedy_inceptionnet_2015}, and ResNet \cite{he_resnet_2016} rely on max or average pooling, whereas spectral subsampling preserves considerably more spatial information per parameter while allowing flexible output dimensionality \cite{rippel_spectral_2015}. Two-dimensional spectral pooling can be based on the DFT \cite{rippel_spectral_2015}, DCT \cite{smith_discrete_2018}, or Hartley transform \cite{zhang_hartley_2018}; among these, the DCT has recurred in fingerprint pipelines \cite{jirachaweng_fingerprint_2007, al-ani_biometrics_2013} and, relying only on real-valued cosine functions rather than complex exponentials, produces output in the real domain $\mathbb{R}$.

We implement our own 2D spectral pooling layer based on the 2D Type-II DCT. Our implementation achieves a significant computational speed-up by decomposing the 2D transform into sequential 1D transforms and pre-computing the spectral transformation bases via an identity-matrix mapping. This reframes the transform as a pure General Matrix Multiply (GEMM) operation, achieving a $50 \times$ speedup on CUDA devices (details in \Cref{app:dct}).

Rather than evaluating the standard separable Type-II DCT summation \cite{ahmed_discrete_1974} (restated in \Cref{app:dct}) directly, our implementation computes the full 2D transform as a single bilinear matrix multiplication:
\begin{equation}
    Y = W_H \, X \, W_W^{\top}
\label{eq:gemm}
\end{equation}
Here $W_H = \mathcal{D}_{1D}(I_H)$ and $W_W = \mathcal{D}_{1D}(I_W)$, where $\mathcal{D}_{1D}$ denotes the one-dimensional Type-II DCT operator, while $I_H \in \mathbb{R}^{H \times H}$ and $I_W \in \mathbb{R}^{W \times W}$ denote identity matrices. The inverse of the 2D Type-II DCT is well defined as the 2D Type-III DCT \cite{ahmed_discrete_1974}; thus the pooled and filtered representation in the frequency domain is transformed back to the spatial domain through the 2D Type-III DCT as the output of the layer.

\begin{figure}[t]
  \centering
  \includegraphics[width=0.46\linewidth]{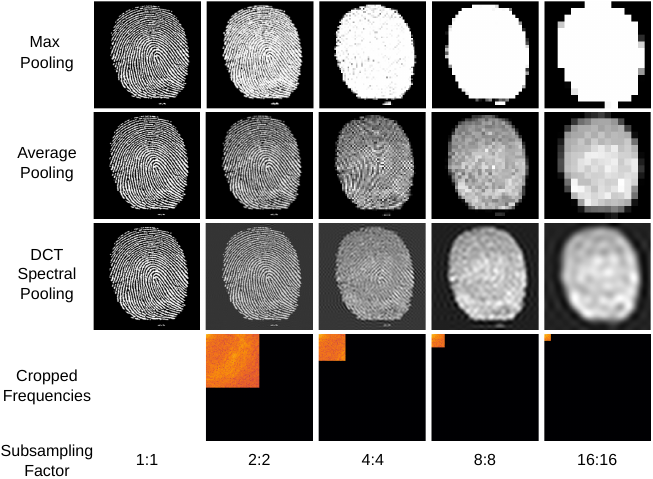}
  \caption{Reconstructions comparing max, average, and DCT spectral pooling on a fingerprint impression. DCT spectral pooling superiorly preserves fine spatial structure and supports arbitrary output dimensions.}
  \label{fig:dct_visualization}
\end{figure}

The layer thus shifts dimensionality reduction into the frequency domain: the Type-II DCT concentrates the structurally significant low-frequency content into the top-left of the frequency matrix; downsampling is a simple crop to the top-left $h_{freq} \times w_{freq}$ block; and the Type-III DCT reconstructs the spatial feature map at a chosen $h_{out} \times w_{out}$. We maintain $h_{freq} \leq h_{out}$ and $w_{freq} \leq w_{out}$, since the converse would discard retained frequencies during the inverse transform and render the crop wasteful. In each convolutional block we use a cropping ratio $h_{freq}/h_{out} \approx w_{freq}/w_{out} \approx 0.85$, avoiding the aggressive reduction of $2{\times}2$ max or average pooling, which discards $75\%$ of the spacial samples per stage. 
\Cref{fig:dct_visualization} illustrates the difference on a fingerprint impression: max and average pooling blur or fragment the ridge structure, while DCT spectral pooling preserves it. Given an input pair $S_a, S_b \in \mathbb{R}^{1 \times 192\times 192}$, the feature extraction module derives feature maps $v_a, v_b \in \mathbb{R}^{512 \times 14\times 14}$.

\subsection{Siamese matching module} \label{subsec: siamese_matchin_module}
While matching is often treated as an independent post-processing step \cite{yi_lift_2016, melekhov_image_2016, oh_deep_2017, han_matchnet_2015, schmidt_self_2016}, the 2D correlation layer of \cite{rocco_convolutional_2017} integrates matching information directly: acting as a normalized cross-correlation $G: \mathbb{R}^{C \times H \times W} \times \mathbb{R}^{C \times H \times W} \rightarrow \mathbb{R}^{(HW) \times H \times W}$, it flattens one feature map to $\mathbb{R}^{C \times N}$ ($N = HW$) and computes, for each spatial feature of the other map, a dot product against all $N$ flattened features. While effective, this breaks all spatial axes of one input while preserving the other. For precise alignment of fine, repeating structure, preserving the spatial hierarchy of \emph{both} images is critical, since the reference and moving images differ fundamentally in their photometric properties. Inspired by the 3D matching approach in Gum-Net \cite{zeng_gum-net_2020}, we therefore propose a symmetric 2D Siamese correlation layer that computes two maps, $C_{ab}$ and $C_{ba}$, processed in parallel.

Given the two outputs of the feature extraction module (\Cref{subsec: feature_extraction_module}), $v_a \in \mathbb{R}^{512 \times 14 \times 14}$ (reference) and $v_b \in \mathbb{R}^{512 \times 14 \times 14}$ (moving), we obtain $C_{ab}, C_{ba} \in \mathbb{R}^{196 \times 14 \times 14}$.

The first correlation map, $C_{ab}$, preserves the spatial coordinates of $v_a$. Each element at channel $k$ (where $k \in \{1, ..., N\}$ indexes a flattened spatial location of $v_b$) and spatial location $(h, w)$ of $v_a$ is the normalized dot product, with $C_{ba}$ defined symmetrically:
\begin{equation}
(C_{ab})_{k,h,w} = \frac{\left\langle v_{a_{:,h,w}},\, v_{b_{:k}} \right\rangle}{\sqrt{\sum \left\langle v_a,\, v_b \right\rangle}}, \qquad
(C_{ba})_{k,h,w} = \frac{\left\langle v_{b_{:,h,w}},\, v_{a_{:k}} \right\rangle}{\sqrt{\sum \left\langle v_b,\, v_a \right\rangle}}
\end{equation}

The correlation maps $C_{ab}, C_{ba}$ are fed into a pseudo-Siamese network of contiguous convolutional layers. Each map is convolved separately, then the branch outputs are concatenated and flattened to predict the dense deformation field.

\subsection{Unsupervised non-linear alignment module} \label{non_linear_alignment}

Unlike the original Gum-Net \cite{zeng_gum-net_2020}, which regresses a rigid 3D transformation suitable for macromolecular structures, our architecture predicts a dense deformation field $\mathbf{\Phi}$. The non-linear alignment module translates the correlation representations from \Cref{subsec: siamese_matchin_module} into a pixel-wise displacement field in two stages: control-point regression and dense flow-field generation.\\
\textbf{Control point regression:} Each correlation branch is reduced by six successive $1024$-channel valid convolutions to spatial size $2 \times 2$ and flattened to $\mathbb{R}^{4096}$; the two branches are concatenated into a joint representation $v_{corr} \in \mathbb{R}^{8192}$ that captures the geometric consensus (full architecture in \Cref{app:arch}). $v_{corr}$ is fed into a multi-layer perceptron with two hidden layers of 2000 ReLU units. The final regressor head projects an output vector $d \in \mathbb{R}^{2G^2}$, where $G$ is the resolution of the sparse control-point grid. We reshape $d$ with the operator $\mathcal{R} : \mathbb{R}^{2G^2} \rightarrow \mathbb{R}^{2 \times G \times G}$, mapping the linear activations onto their spatial coordinates on the control-point grid to obtain $\Delta {\mathbf{P}} = (\Delta{x}, \Delta{y})$.\\
\textbf{Dense flow field generation:} To generate a smooth, topology-preserving dense deformation field, we upsample the coarse control points to the input resolution $(H, W)$ using bicubic interpolation:
\begin{equation}
\mathbf{\Phi} = \text{Upsample}_{bicubic}(\Delta \mathbf{P})
\end{equation}
where $\mathbf{\Phi} \in \mathbb{R}^{H \times W \times 2}$. The final aligned image $S_{\hat{b}}$ is generated by spatially resampling the moving image $S_b$ according to $\mathbf{\Phi}$:
\begin{equation}
S_{\hat{b}} = S_b \circ \mathbf{\Phi}
\end{equation}
This parameterization is, structurally, a free-form deformation model: any field in the network's range is a bicubic lift of a coarse lattice, so smoothness is enforced \emph{architecturally}, as high-frequency or folding fields are inexpressible, and is only refined, not created, by the regularizer below. It is the learned analogue of classical spline-based elastic registration \cite{bazen_fingerprint_2003, ross_warping_2006}, with correspondence extraction and per-pair optimization amortized into one forward pass.

\subsection{Training objective} \label{subsec:objective}
To quantify pixel-wise alignment between the reference $S_a$ and the warped image $S_{\hat{b}}$, we use the soft Dice loss \cite{wang_dice_2023}. Let $N$ be the number of pixels in the flattened images and $k$ index a batch of size $B$:
\begin{equation}
\mathcal{L}_{\text{dice}}=\frac{1}{B}\sum_{k=1}^{B}\left(1-\frac{2\sum_{i=1}^{N}S_{\hat{b},k}^{(i)}S_{a,k}^{(i)}+\epsilon}{\sum_{i=1}^{N}S_{\hat{b},k}^{(i)}+\sum_{i=1}^{N}S_{a,k}^{(i)}+\epsilon}\right)
\label{eq:dice}
\end{equation}
On near-binary ridge imagery, this is directly interpretable as the fraction of structure brought into coincidence. Let $V$ denote the predicted control-point grid; we apply a first-order smoothness penalty
\begin{equation}
\mathcal{L}_{\text{smooth}}=\mathbb{E}\left[\left\|\nabla_{x}V\right\|^{2}+\left\|\nabla_{y}V\right\|^{2}\right]
\label{eq:smooth}
\end{equation}
and train with the combined objective
\begin{equation}
\mathcal{L}_{\text{total}}=\mathcal{L}_{\text{dice}}+\lambda_{\text{reg}}\mathcal{L}_{\text{smooth}}.
\label{eq:total}
\end{equation}
No deformation ground truth, correspondence annotation, or identity label is used at any stage.

\section{Experiments}
\label{sec:experiments}
We train Gum-Net entirely on synthetic data and evaluate on synthetic as well as real-world data. Unlike \cite{zeng_gum-net_2020}, we perform no fine-tuning or adaptation on real-world datasets: the model is evaluated \emph{zero-shot} on real benchmarks, so all real-data results directly measure synthetic-to-real transfer of the learned registration function.

\subsection{Datasets}
\paragraph{Real datasets.}
As real-world benchmarks providing organically generated elastic deformation, we evaluate on the FVC2004 benchmark \cite{maio_fvc2004_2004} (110 fingers, 8 impressions each across 4 databases) and the Tsinghua Distorted Fingerprint Dataset (320 pairs of undistorted and distorted impressions) \cite{si_detection_2015}. We use FVC2004 to quantify pre-matching structural alignment improvements via the Pearson Correlation Coefficient (PCC). Conversely, the Tsinghua dataset features severe real-world torque and complex skin distortion, and we use it exclusively to stress-test downstream minutiae matching behavior.

\paragraph{Simulated datasets.}
We use Anguli \cite{ansari_generation_2011}, a free and open-source synthetic fingerprint generator developed at the Indian Institute of Science, Bengaluru, to generate 73{,}500 fingerprints: 2{,}100 distinct impressions per noise level per class, across five signal-to-noise ratios and all fingerprint classes of the Henry classification \cite{henry_classification_1928}. The generator first produces a master template image, then casts impressions from that template (\Cref{fig:templates,fig:dataset} in the appendix show the templates and corpus diversity). Anguli is a \emph{procedural} generator: it synthesizes orientation fields from mathematical models of singular points and grows ridge patterns by iterative filtering \cite{ansari_generation_2011, cappelli_sfinge_2004}. It is not a learned generative model fitted to real fingerprints, so synthetic training involves no real biometric data at any stage; real datasets are used strictly for evaluation. Beyond scale, this yields unlimited reference--moving training pairs with correspondence guaranteed by construction.

\subsection{Training}
\begin{figure}[t]
  \centering
  \includegraphics[width=0.64\linewidth]{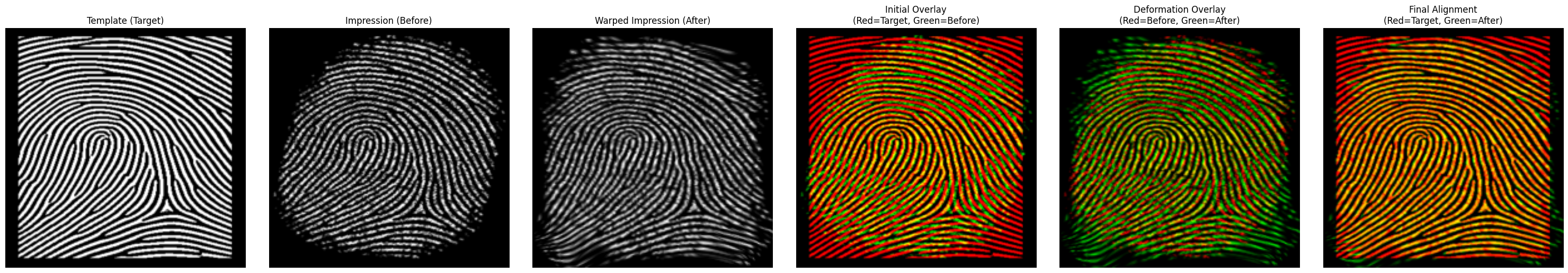}
  \caption{Inference plot obtained from test data. This is an impression the model has never seen during training or validation.}
  \label{fig:infplot}
\end{figure}
The models are implemented in PyTorch \cite{paszke_pytorch_2019}. All inputs have size $192^2$, though the DCT spectral pooling layers make the input size arbitrary in principle. We train a single Gum-Net model on the entire synthetic dataset (seven fingerprint classes, five noise levels) with no labels at any level, following a curriculum \cite{bengio_curriculum_2009}: training begins on the least noisy data and gradually introduces higher noise, 100 epochs per level with 14{,}700 examples each, concluding after 500 epochs. A control-point grid resolution $G=8$ is used throughout, with the objective $\mathcal{L}_{\text{total}}$ of \Cref{eq:total}.
\Cref{fig:infplot} shows a representative registration of an unseen test impression; further examples across all finger classes are given in \Cref{fig:infplotfull} (\Cref{app:inference}).

\subsection{Evaluation}
\begin{table}[t]
\centering
\caption{Downstream metrics, synthetic test set. Gum-Net improves genuine matching ($\mu_{gen}$) and decidability ($d'$) while holding imposter scores ($\mu_{imp}$) stable, outperforming the unaligned baseline and the AP/MP pooling variants.}
\label{tab:downstream_metrics}
\small
\begin{tabular}{lcccccc}
\toprule
\textbf{Model} & $\mu_{gen}$ $\uparrow$ & $\mu_{imp}$ $\downarrow$ & $d'$ $\uparrow$ & $\Delta\mu_{gen}$ $\uparrow$ & $\Delta\mu_{imp}$ $\downarrow$ & $\Delta d'$ $\uparrow$ \\
\midrule
Baseline & 153.25 & 10.06 & 4.0610 & 0 & 0 & 0 \\
GumNet-AP & 152.65 & 10.30 & 3.7788 & -0.60 & 0.24 & -0.2823 \\
GumNet-MP & 123.75 & 9.39 & 3.4025 & -29.49 & -0.67 & -0.6585 \\
\textbf{GumNet} & \textbf{170.39} & \textbf{10.25} & \textbf{4.8739} & \textbf{17.14} & \textbf{0.19} & \textbf{0.8128} \\
\bottomrule
\end{tabular}
\end{table}

\paragraph{Evaluation of registration quality.}
To quantify the performance of the predicted dense deformation field, we measure pixel-level structural correspondence between the reference $S_a$ and the rectified image $S_{\hat{b}}$ using the Pearson Correlation Coefficient (PCC):
\begin{equation}
\rho_{S_a, S_{\hat{b}}} = \frac{\sum_{i=1}^{N}(S_{a, i}-\bar{S}_a)(S_{\hat{b}, i}-\bar{S}_{\hat{b}})}{\sqrt{\sum_{i=1}^{N}(S_{a, i}-\bar{S}_a)^2 \sum_{i=1}^{N}(S_{\hat{b}, i}-\bar{S}_{\hat{b}})^2}}
\end{equation}
where $N$ is the total pixel count and $\bar{S}_a$, $\bar{S}_{\hat{b}}$ denote mean intensities. On near-binary ridge imagery, intensity correlation directly measures structural coincidence of ridge patterns \cite{lan_pre-registration_2020, press_numerical_2007}; the improvement from $\rho(S_a, S_b)$ to $\rho(S_a, S_{\hat{b}})$ therefore evaluates unwarping capability independently of any downstream matching heuristic.

\begin{table}[t]
\centering
\caption{Pearson correlation coefficient increase in \% on the FVC2004 datasets DB1--DB3 and our synthetic dataset.}
\label{tab:PCC results}
\small
\begin{tabular}{lcccc}
\toprule
\textbf{Model} & \textbf{DB1} & \textbf{DB2} & \textbf{DB3} & \textbf{Synthetic (Ours)} \\
\midrule
No alignment & +0.0 & +0.0 & +0.0 & +0.0\\
ORB alignment & -5.0 & -4.6 & -7.4 & -3.0\\
Gum-Net AP   & +8.7 & +12.6 & +10.7 & +6.7\\
\textbf{Gum-Net} & \textbf{+11.2} & \textbf{+15.7} & \textbf{+11.3} & \textbf{+10.5}\\
\bottomrule
\end{tabular}
\end{table}

We compare against an ORB baseline \cite{Rublee_ORB_2011}, a feature-matching approach that strictly models linear (rigid/affine) transformations. As shown in \Cref{tab:PCC results}, Gum-Net substantially outperforms ORB, achieving up to a 15.7\% PCC increase on FVC2004 intra-class pairs with no real-data training. ORB in fact \emph{degrades} baseline alignment on every dataset, which indicates that the deformation in these pairs lies outside the linear transformation class.

\paragraph{Topological validation via an independent minutiae matcher.}
A registration method can inflate similarity metrics by hallucinating structure; the real question is whether the predicted field restores \emph{true} topology. Our testbed supplies an external oracle for exactly this question: minutiae are precise point landmarks consumed by matchers entirely independent of our training objective, so a topology-restoring warp raises genuine match scores, a hallucinating warp is penalized, and, since no warp should manufacture similarity between \emph{different} fingers, imposter scores should remain unchanged. Concretely, we evaluate with the NIST Bozorth3 \cite{watson_nbis_2007} matcher: distorted images are unwarped via the predicted fields, minutiae are extracted from the aligned images, and matching scores form genuine and imposter distributions whose separability is summarized by the decidability index $d' = |\mu_{gen} - \mu_{imp}| \,/\, \sqrt{\tfrac{1}{2}(\sigma^2_{gen} + \sigma^2_{imp})}$, where $\mu$ and $\sigma^2$ denote the means and variances of the genuine and imposter score distributions.

\Cref{tab:downstream_metrics} reports these metrics on the synthetic test set and \Cref{tab:tsinghua_matching} on the real-world Tsinghua dataset. Across both domains, Gum-Net substantially increases genuine-pair scores ($\Delta \mu_{gen} \gg 0$) with negligible perturbation of imposter pairs ($\Delta \mu_{imp} \approx 0$), outperforming the unaligned baseline and the AP/MP variants. This asymmetry is what one would expect if the warps recover real structure: alignment increases similarity only where genuine correspondence exists, indicating that the predicted fields do not fabricate matches. These are distribution-level summaries rather than operating characteristics; full ROC/DET behavior of a complete recognition pipeline is beyond our pre-matching scope and left to future work.

\begin{table}[t]
\centering
\caption{Downstream matching behavior on the real-world Tsinghua dataset.}
\label{tab:tsinghua_matching}
\small
\begin{tabular}{lcccc}
\toprule
\textbf{Alignment Method} & \textbf{$\mu_{gen}$} & \textbf{$\Delta \mu_{gen}$} & \textbf{$d'$} & \textbf{$\Delta d'$} \\
\midrule
Baseline (No Alignment) & 40.88 & -- & 1.4111 & -- \\
Gum-Net 2D (Max Pooling) & 42.90 & +2.01 & 1.4227 & +0.0116 \\
Gum-Net 2D (Avg Pooling) & 44.03 & +3.15 & 1.4749 & +0.0638 \\
\textbf{Gum-Net 2D (DCT)} & \textbf{43.97} & \textbf{+3.09} & \textbf{1.4775} & \textbf{+0.0664} \\
\bottomrule
\end{tabular}
\end{table}

Two observations deserve comment. First, on synthetic data the max- and average-pooling variants do not merely underperform; they \emph{reduce} decidability relative to no alignment at all: alignment quality, not alignment presence, drives the downstream benefit, and a poorly estimated warp is worse than none. Second, on Tsinghua the average-pooling variant attains a marginally higher genuine mean than DCT (44.03 vs.\ 43.97), while DCT retains the best decidability. We hypothesize that Tsinghua's enhanced, binarized impressions reduce the fine-detail advantage spectral pooling provides on raw ridge imagery; the unambiguous synthetic ordering isolates the effect on unprocessed structure. Characterizing this preprocessing--pooling interaction remains open.

\section{Conclusion}
We introduced 2D Gum-Net, an unsupervised framework for dense non-rigid 2D registration that transfers 3D Gum-Net \cite{zeng_gum-net_2020} from cryo-ET to two dimensions and replaces its rigid head with a Dense Spatial Transformer, while DCT spectral pooling and control-point parameterization preserve the fine structure on which correspondence depends. Trained only on synthetic fingerprints and evaluated zero-shot, it improves structural correlation by 10.5\% (synthetic) and up to 15.7\% (FVC2004), and on Tsinghua raises genuine match scores while leaving imposter scores essentially unchanged. Evaluation is limited to one domain and to distribution-level summaries; future work includes other imagery where fine quasi-periodic structure meets elastic deformation, operational metrics on larger datasets, and coupling with fixed-length embeddings \cite{engelsma_deepprint_2019} for retrieval at scale.

{
\small
\bibliographystyle{unsrtnat}
\bibliography{main}
}

\appendix
\section{Detailed architecture of Gum-Net variants}
\label{app:arch}
\begin{figure}[htbp]
  \centering
  \includegraphics[width=0.42\linewidth]{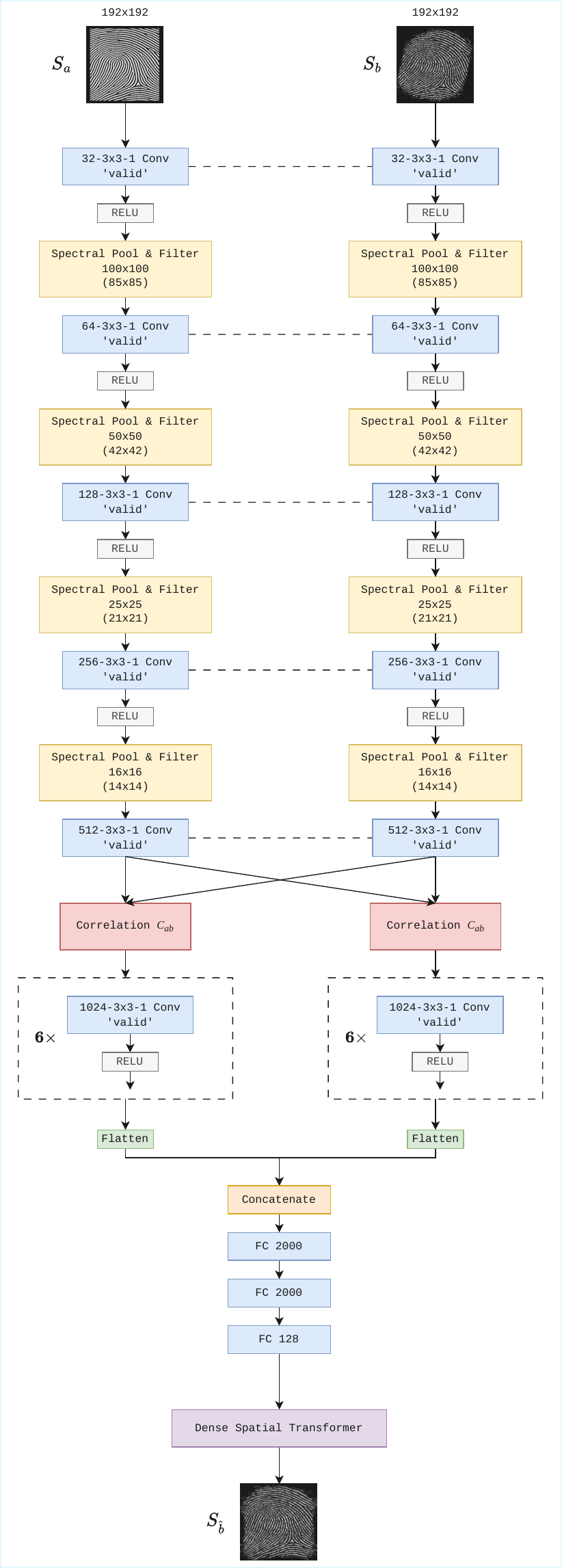}
  \caption{Detailed architecture of Gum-Net.}
  \label{fig:detarch}
\end{figure}

\begin{figure}[htbp]
  \centering
  \includegraphics[width=0.46\linewidth]{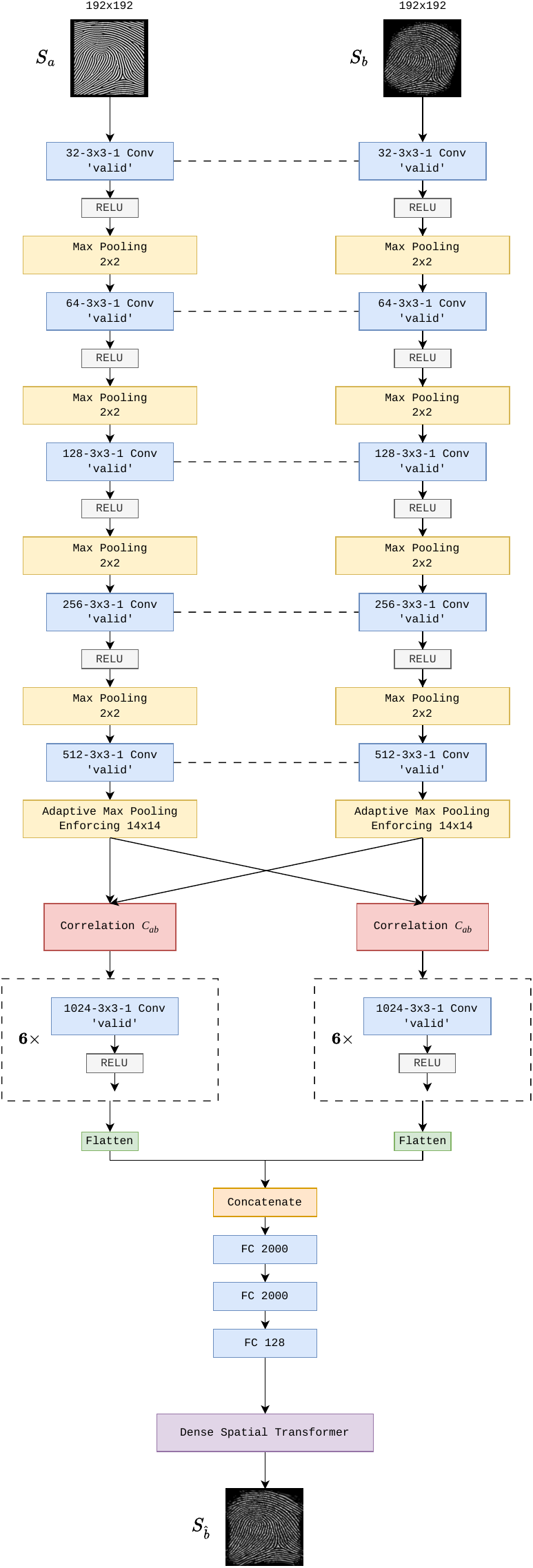}
  \caption{Detailed architecture of Gum-Net MP.}
  \label{fig:detarchmp}
\end{figure}

\begin{figure}[htbp]
  \centering
  \includegraphics[width=0.46\linewidth]{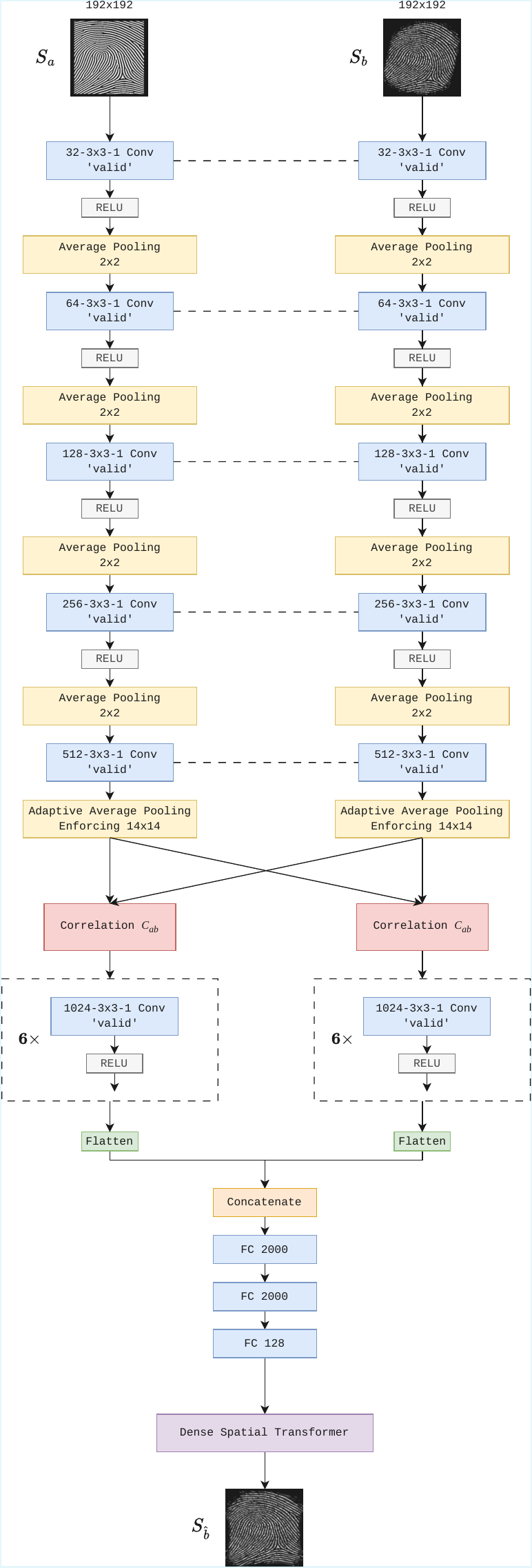}
  \caption{Detailed architecture of Gum-Net AP.}
  \label{fig:detarchap}
\end{figure}

To verify that DCT spectral pooling and filtering layers offer superior performance compared to max and average pooling, we built identical architectures differing only in the pooling operation, named Gum-Net MP and Gum-Net AP. \Cref{fig:detarch} shows the detailed architecture for Gum-Net with input size $192\times192$ for both $S_a$ and $S_b$; \Cref{fig:detarchmp,fig:detarchap} show the detailed architectures for Gum-Net MP and Gum-Net AP respectively. Gum-Net AP and Gum-Net MP change only the pooling within the feature extraction module (\Cref{subsec: feature_extraction_module}), keeping the rest of the architecture exactly the same.

\section{Dataset diversity and inference details}
\label{app:inference}
\begin{figure}[htbp]
  \centering
  \includegraphics[width=0.6\linewidth]{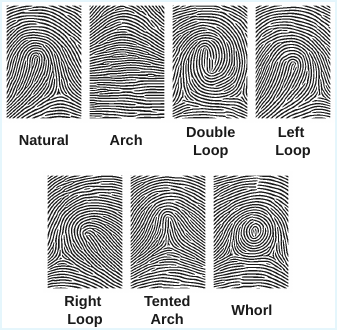}
  \caption{Templates used by Anguli to generate the fingerprint impressions.}
  \label{fig:templates}
\end{figure}

\begin{figure}[htbp]
  \centering
  \includegraphics[width=0.8\linewidth]{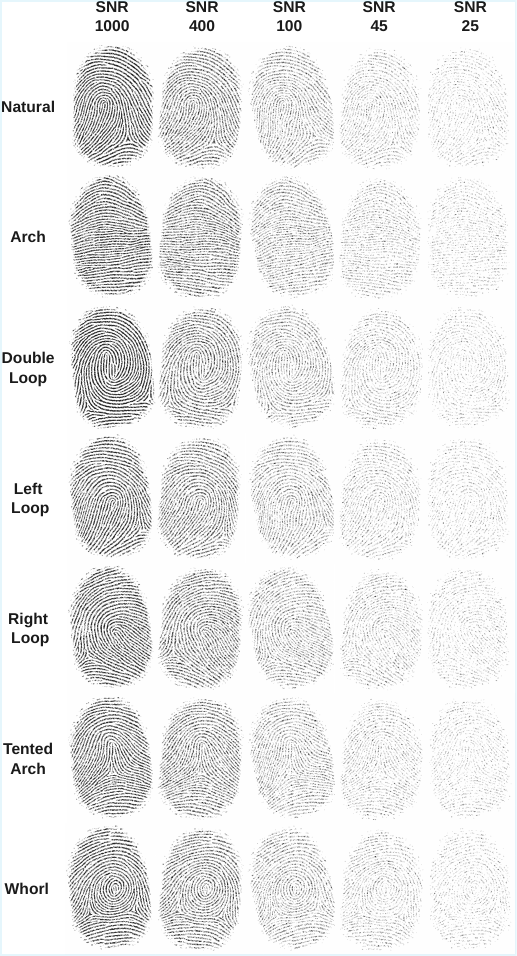}
  \caption{Fingerprint impressions across seven classifications and five signal-to-noise ratios.}
  \label{fig:dataset}
\end{figure}

\begin{figure}[htbp]
  \centering
  \includegraphics[width=1.0\linewidth]{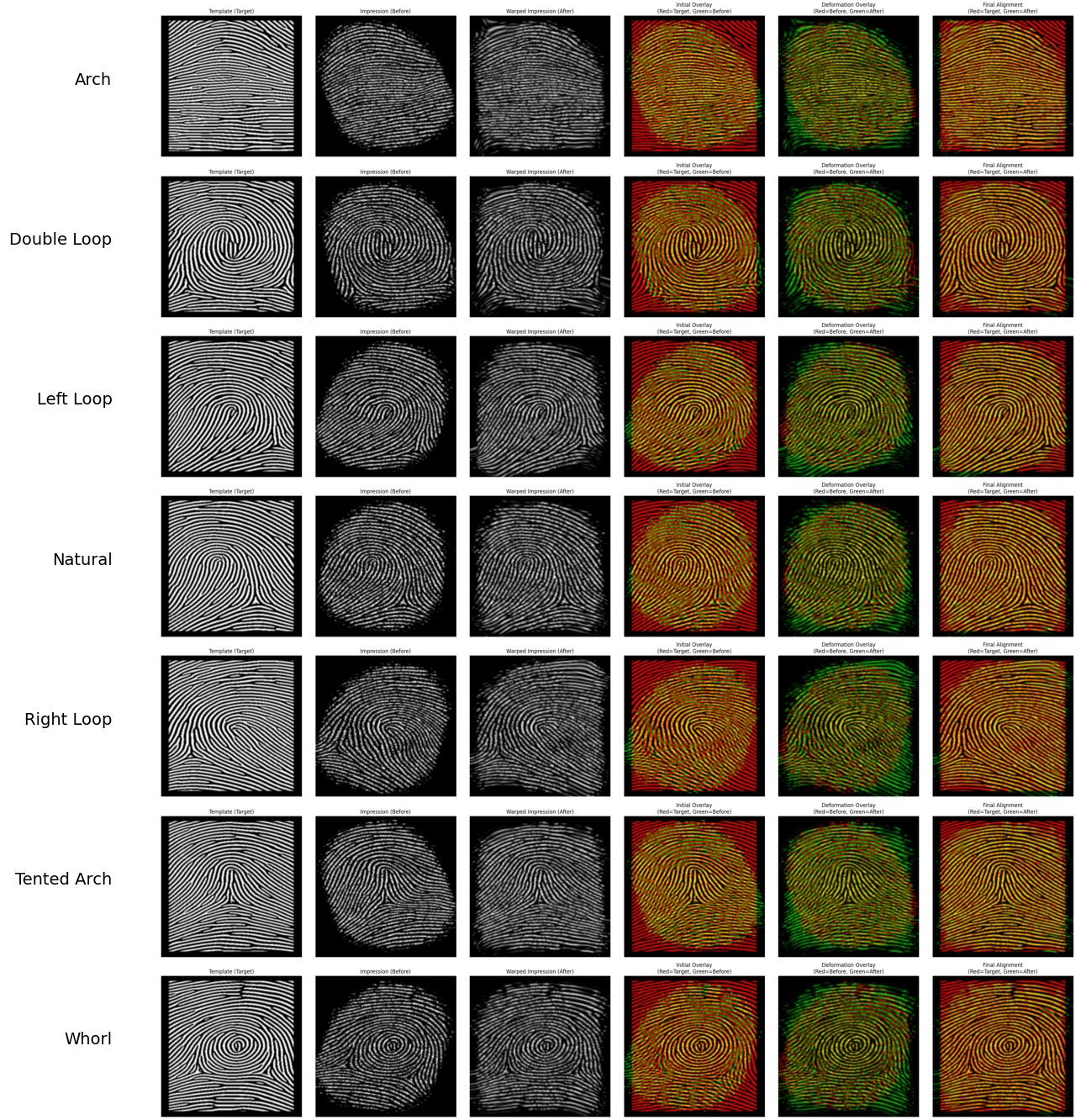}
  \caption{Inference plots across all finger types.}
  \label{fig:infplotfull}
\end{figure}

\Cref{fig:templates} shows the master templates used by Anguli, and \Cref{fig:dataset} illustrates the diversity of our synthetic corpus across the seven fingerprint classes and five signal-to-noise ratios. \Cref{fig:infplotfull} shows inference plots across all finger types; the model has not seen these reference--moving pairs during training.

\clearpage
\section{Details on the DCT optimization}
\label{app:dct}
The 2D DCT spectral pooling layer is implemented as a sequence of frozen linear transformations to exploit highly optimized General Matrix Multiply (GEMM) routines on CUDA devices. For a 1D signal of length $N$, the Type-II DCT basis matrix $\mathbf{W} \in \mathbb{R}^{N \times N}$ is precomputed by applying the DCT operator to the identity matrix $\mathbf{I}_N$:
\begin{equation}
    \mathbf{W} = \text{DCT-II}(\mathbf{I}_N)
\end{equation}
such that the transform $\mathbf{y} = \mathbf{x}\mathbf{W}^\top$ is equivalent to a standard 1D DCT-II applied to the row vector $\mathbf{x}$. This matrix is computed once at initialization and stored as a frozen (non-trainable) weight in a linear layer, reducing repeated transform computation to a single matrix multiplication at inference time.

The 2D Type-II DCT of a feature map $\mathbf{X} \in \mathbb{R}^{C \times H \times W}$ is defined as \cite{ahmed_discrete_1974}
\begin{equation}
Y_{c, u, v} = \alpha_u \alpha_v \sum_{x=0}^{H-1} \sum_{y=0}^{W-1} X_{c, x, y} \cos\left(\frac{\pi u (2x+1)}{2H}\right) \cos\left(\frac{\pi v (2y+1)}{2W}\right)
\label{eq:dctfull}
\end{equation}
with $\alpha_u = \sqrt{1/H}$ for $u=0$ and $\sqrt{2/H}$ for $u>0$, $\forall u \in \{0, ..., H-1\}$, and $\alpha_v$ defined analogously. Its separability allows the transform to be decomposed into two sequential 1D operations:
\begin{equation}
    \mathbf{Y} = \mathbf{W}_H \, \mathbf{X} \, \mathbf{W}_W^\top
\end{equation}
where $\mathbf{W}_H = \text{DCT-II}(\mathbf{I}_H) \in \mathbb{R}^{H \times H}$ and $\mathbf{W}_W = \text{DCT-II}(\mathbf{I}_W) \in \mathbb{R}^{W \times W}$ are the precomputed basis matrices for the height and width dimensions respectively. In practice, this is computed by first applying $\mathbf{W}_W$ along the width axis, transposing, applying $\mathbf{W}_H$ along the (now leading) spatial axis, and transposing back, matching \Cref{eq:gemm} in the main paper.

Following the forward 2D DCT, a binary low-pass mask $\mathbf{M} \in \{0,1\}^{H \times W}$ is applied element-wise to the frequency representation:
\begin{equation}
    \tilde{\mathbf{Y}} = \mathbf{Y} \odot \mathbf{M}, \qquad \mathbf{M}_{u,v} =
    \begin{cases}
        1 & \text{if } u < h_{\text{freq}} \text{ and } v < w_{\text{freq}} \\
        0 & \text{otherwise}
    \end{cases}
\end{equation}
retaining only the top-left $h_{\text{freq}} \times w_{\text{freq}}$ block of low-frequency coefficients. The masked representation is then cropped to $h_{\text{out}} \times w_{\text{out}}$ and reconstructed via the inverse transform, where the 2D Type-III DCT (the inverse of Type-II) is applied separably using analogous precomputed basis matrices $\mathbf{W}_{h_{\text{out}}}^{-1}$ and $\mathbf{W}_{w_{\text{out}}}^{-1}$:
\begin{equation}
    \hat{\mathbf{X}} = \mathbf{W}_{h_{\text{out}}}^{-1} \, \tilde{\mathbf{Y}}_{[:h_{\text{out}}, :w_{\text{out}}]} \, \left(\mathbf{W}_{w_{\text{out}}}^{-1}\right)^\top
\end{equation}
The constraint $h_{\text{freq}} \leq h_{\text{out}}$ and $w_{\text{freq}} \leq w_{\text{out}}$ is enforced to ensure that no retained frequency coefficients are discarded during the inverse transform, which would render the masking step computationally redundant. This GEMM-based implementation achieves approximately a $50\times$ speedup over a naive iterative DCT on CUDA devices, as matrix multiplication is among the most heavily optimized parallel primitives available on modern GPU hardware.

\begin{figure}[htbp]
  \centering
  \includegraphics[width=0.75\linewidth]{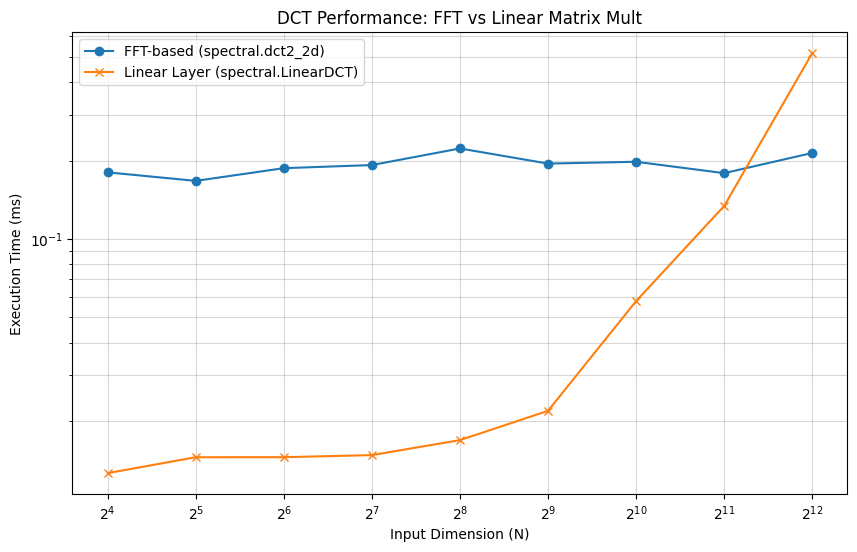}
  \caption{Time comparison.}
  \label{fig:dct_tradeoff_time}

  \vspace{1em}

  \includegraphics[width=0.75\linewidth]{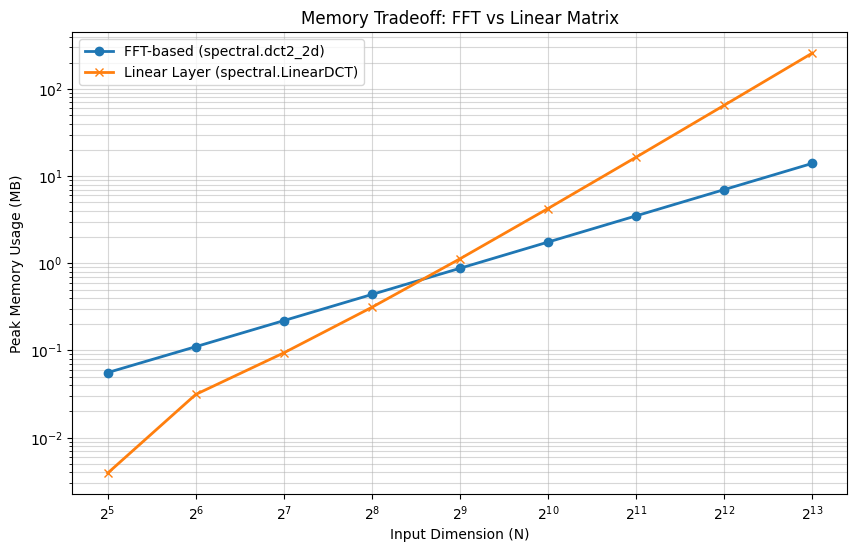}
  \caption{Space comparison.}
  \label{fig:dct_tradeoff_space}
\end{figure}

\Cref{fig:dct_tradeoff_time,fig:dct_tradeoff_space} show our speedup and memory efficiency compared to a traditional implementation of the DCT in PyTorch, which uses the built-in FFT suite to extract DCT coefficients.

All code for this project is available at \url{https://github.com/caffeinekeyboard/2DGumNet}.

\end{document}